\documentclass{llncs}
\usepackage[a4paper,margin=1.5in]{geometry}
\usepackage{polyglossia}
\setdefaultlanguage[variant=uk]{english}
\usepackage{csquotes}
\usepackage{diagbox}
\usepackage{mathtools}
\usepackage{pgfplots}
\usepackage{tikz}
\usepackage{natbib}
\setcitestyle{authoryear}
\usepackage{graphicx}
\graphicspath{{Fig/}}
\usepackage{algorithm}
\usepackage{algpseudocode}
\usepackage[hidelinks,bookmarks=true]{hyperref}
\usepackage{bookmark}
\usepackage{libertine}
\usepackage[T1]{fontenc}

\title{Modelling Intertextuality with N-gram Embeddings}
\author{Yi Xing}
\institute{Centre for Perceptual and Interactive Intelligence,\\
The Chinese University of Hong Kong\thanks{Work done while at Nankai University}}

\begin{document}

\maketitle
\begin{abstract}
	Intertextuality is a central tenet in literary studies. It refers to the intricate links between literary texts that are created by various types of references. This paper proposes a new quantitative model of intertextuality to enable scalable analysis and network-based insights: perform pairwise comparisons of the embeddings of n-grams from two texts and average their results as the overall intertextuality. Validation on four texts with known degrees of intertextuality, alongside a scalability test on 267 diverse texts, demonstrates the method's effectiveness and efficiency. Network analysis further reveals centrality and community structures, affirming the approach's success in capturing and quantifying intertextual relationships.
\end{abstract}


\section{Introduction}

\begin{figure}[h]
	\centering
	\includegraphics[width=0.6\textwidth]{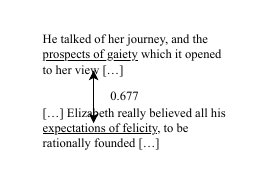}
	\caption{An intertextual link between Frances Burney's \textit{Cecilia} and Jane Austen's \textit{Pride and Prejudice} established by semantically similar trigrams}
\end{figure}

Intertextuality, the allusive relationship between literary texts, is a fundamental concept in literary studies. It is the idea that texts are not isolated entities, but are interconnected through a network of references, allusions, and influences. Intertextuality is a key aspect of both literary creativity and interpretation, and it has been a popular research topic since it was put forward by the French semiotician Julia Kristeva in the 1960s \citep{kristeva_word_2024,kristeva_poesie_1968}.

Traditionally, the analysis of intertextuality has been a qualitative and interpretative endeavour, relying on close reading and critical judgement, and focusing only on a small number of texts. With the advent of digital humanities, it has become possible to study intertextuality quantitatively on a much larger scale, using computational methods to analyse entire corpora of texts. Substantial progress has been made over the past few decades, and a number of ready-to-use research tools have been developed. Among existing methods, two categories, based on n-gram alignment and embedding respectively, stand out.

The first method is exemplified by the tool TextPAIR \citep{gladstone_textpair_2009}, which works by extracting all overlapping n-grams, or `shingles' \citep{broder_resemblance_1997}, from one text, then aligning them with n-grams from another text. It has been successfully used to analyse the complex intertextual interactions between French Enlightenment philosophical texts \citep{allen_plundering_2010,fedchenko_a_2024}, among other applications.

Another tool that finds similar passages by exploiting the appearance of overlapping n-grams is Passim \citep{smith_infectious_2013}, which is able to handle large-scale corpora with effective indexing, hashing and pruning, and has been used to detect intertextuality within newspapers \citep{smith_computational_2015,during_impresso_2023} and patristic sermons \citep{denis_pseudo-augustinian_2024}.

Other examples include the Tesserae project, which focuses on the intertextual analysis of Latin poetry. Its original version considers two six-word passages to constitute a parallel if they share two or more words \citep{coffee_intertextuality_2012,coffee_tesserae_2013}, and it has evolved by including n-gram matching \citep{forstall_modeling_2015} and other methods. Likewise, the R package \textit{textreuse} implements n-gram-based pairwise comparison that can be optimised with the use of MinHash and locality-sensitive hashing \citep{li_textreuse_2024}.

The second method avoids the problem of exact lexical literalness of n-gram matching with semantically enriched word embeddings such as Word2Vec \citep{mikolov_efficient_2013} and FastText \citep{bojanowski_enriching_2017}. \citet{zhang_continuous_2014} obtain word vectors with NNLM \citep{bengio_neural_2003} and Word2Vec, then aggregate the vectors of all the words in a text into a fixed-length vector to detect local text reuse in social media posts; \citet{barbu_intertextuality_2017} perform pairwise comparisons of the Word2Vec embeddings of words from two documents to study literary intertextuality; \citet{liebl_shakespeare_2020} align text sequences with FastText and wnet2vec embeddings to detect Shakespearean quotations; \citet{burns_profiling_2021} detect intertextuality in Latin literature by computing similarity between two bigrams from different texts: all four possible combinations' similarity is calculated, then the bigrams' similarity is obtained by averaging the highest similarity score of the four and the remaining two's.

Apart from word embeddings, sentence and document embeddings have also been used. \citet{barre_latent_2024} uses BGE M3 embeddings \citep{chen_m3-embedding_2024} to find similar passages; \citet{mcgovern_characterizing_2025} characterise intertextuality within translations with sentence embeddings.

However, much of the existing literature aims at locating local passages that constitute intertextual parallels, instead of obtaining a global quantitative intertextuality score.\footnote{Literal n-gram matching can also yield a quantified intertextuality score, for example, by calculating the Jaccard coefficient, but it still suffers from literalness.} This paper proposes a novel method by combining the above ideas---perform pairwise comparisons of the embeddings of the n-grams extracted from two documents, and then average the similarities as a measure of intertextuality between them. As it shall be demonstrated below, this distributional-semantics-based \citep{firth1957synopsis} approach can effectively quantify the degree of intertextuality between texts and can be implemented efficiently.

The contributions of this paper are threefold. First and foremost, it demonstrates the potential of using n-gram embeddings to analyse the intertextuality between any two texts, then constructing a graph where nodes represent texts and edges the degree of intertextuality, which enables us to perform network analysis on it. A validation study and subsequent network analysis both show the effectiveness of the proposed model of intertextuality. Secondly, a simple, computationally efficient, rule-based historical text normalisation pipeline for English was devised. This pipeline was applied to the Oxford Text Archive \citep{ota} to train a diachronic word embedding model covering the 16th to early 20th centuries. Thirdly, an innovative method is presented for evaluating a word embedding model's ability to represent synonymity and antonymy.

\section{Modelling intertextuality}

It is well known that the concept of intertextuality can be and has been defined variously, and here the proposed model is based on the original definition put forward by \citet{kristeva_poesie_1968}. She stated that one poetic signified refers to other signifieds, so that multiple discourses are incorporated into one text, and this superset, or intertextual space, enriches the meaning of the original poetic statement. While this discussion mainly concerns poetry, the concept of intertextuality is equally applicable to other genres. Although there is a myriad of ways in which one signified can reference another, \emph{semantic relatedness} is the easiest to encapsulate in terms of computability. Texts can be embedded into a continuous vector space based on co-occurrence data, and in this space, semantically related texts are close to each other, and their relatedness can be readily computed with established measures like cosine similarity or Euclidean distance.

More recently, \citet{allen_intertextuality_2022} analysed a passage from Balzac's \textit{Sarrasine} and paid special attention to the word `sensibility' (`finesse de sentiments' in Balzac's original French). It is worth noting that single words like `finesse' and `sentiment' are too general to generate strong intertextuality, whereas the n-gram combining them is more specific and potent, and as Allen argues, it could have `a potentially vast array of cultural and literary resonances' and conjure up diverse images from other texts, creating a multitude of intertextual relationships.

Therefore, intertextuality can be modelled through a `bottom-up' approach \citep{forstall_quantitative_2019} as follows: given two texts A and B, for each n-gram in A, compute how strongly it references each n-gram from B by calculating the similarity between their vector representations, and the overall intertextuality is obtained by averaging the results of all pairwise comparisons. This method starts by identifying localised semantic links---the \emph{loci similes} in traditional terms---rather than attempting a top-down, holistic comparison of abstract themes. The underlying principle is that a text's overarching themes and style emerge from the aggregation of its local semantic content \citep{iser_grasping_1980}. This idea, that global patterns can be inferred from local semantics, is a foundational concept in computational linguistics, demonstrated in fields like topic modelling \citep{blei_probabilistic_2012}.

It is noteworthy that the majority of n-grams in a text are not powerfully evocative, and that most pairwise comparisons are meaningless. As an illustration, there are many texts reusing the rose as a sign of short-lived beauty to be enjoyed, for example, Robert Herrick's `Gather ye rosebuds while ye may,' Tasso's `Let's pluck the rosebud while we may' (\textit{The Liberation of Jerusalem}, 16.15), and a line of Latin poetry `Maiden, gather roses, while blossom is fresh and youth is fresh' (\textit{De Rosis Nascentibus}, 49; \textit{Appendix Vergiliana}, sometimes attributed to Ausonius). While it is indisputable that these lines are intertextually linked, some n-gram pairs do not have meaningful links at all, such as `gather ye rosebuds' and `while we may,' and including them in the calculation of the overall intertextuality is nonsensical. Indeed, their mathematical representations are usually orthogonal. Consequently, a threshold is in order when averaging the pairwise comparison results, and it is an adjustable hyperparameter: a low threshold might help capture distant thematic resonances but also introduce noise, while a high threshold might only capture near-quotations.

In summary, the intertextuality between two texts is determined by the equation below:

\begin{equation}
	\mathrm{Intertextuality}(A, B) = \frac{1}{|S|} \sum_{s \in S} s,
\end{equation}
where $A$ and $B$ are the sets of the vector representations of the n-grams extracted from the texts, and $S$ is the set of filtered pairwise comparison results:

\begin{equation}
	S = \{\mathrm{cosine}(\mathbf{a}, \mathbf{b}) | \mathbf{a} \in A \land \mathbf{b} \in B \land |\mathrm{cosine}(\mathbf{a}, \mathbf{b})| > \tau\}.
\end{equation}

Here cosine similarity is used in favour of, for instance, Euclidean distance. N-grams can be directly embedded as a whole with a modern sentence embedding model, though they can also be represented by the average of the vector representations of its constituent words, and, since we are using the cosine, the sum suffices.\footnote{Averaging them effectively discards the word order, making the method theoretically more akin to bag-of-words than n-gram. Therefore, a more accurate name might be `bag-of-words-in-an-n-gram-window,' but for simplicity's sake, we will continue using `n-gram embedding.'} Additionally, earlier word embedding models like Word2Vec and GloVe \citep{pennington_glove_2014} are static, making them much less computationally expensive than sentence embedding models.

The choice of n-grams is also an interesting question. TextPAIR removes all function words from the original text before performing n-gram matching; \citet{manjavacas_statistical_2020} only kept adjectives, adverbs, nouns and verbs when trying to detect intertextuality with topic modelling. However, it is found that keeping pronouns and prepositions can make the comparison more structure-aware. That is, if we take the preposition `of' into account, n-grams in the form of `X of Y' will be more similar to each other than those without the preposition. Therefore, we do not remove such words when calculating intertextuality.

\begin{example}
	Consider the first two lines on budding roses that are listed above, which yield the following n-grams respectively (let $n = 3$):

	\begin{enumerate}
		\item gather ye rosebuds, ye rosebuds while, rosebuds while ye, while ye may
		\item let's pluck the, pluck the rosebud, the rosebud while, rosebud while we, while we may
	\end{enumerate}

	To compute the intertextuality between the first and second lines, we compare each n-gram in the first line individually against the ones from the second and obtain $4 \times 5$ pairwise comparison results. Then we filter out n-gram pairs that are almost orthogonal and average the rest as the final intertextuality score.
\end{example}

\section{Dataset and embedding model}

There are many publicly available word embedding models, such as GloVe models trained on the English Wikipedia and Common Crawl \citep{pennington_glove_2014} and Word2Vec models trained on the Google Books N-gram Corpus (consisting of English books from 1800s to 1990s) and the Corpus of Historical American English \citep{hamilton_diachronic_2016}. However, these models fall short of our needs---they are all restricted to one specific time period or region, and the vocabulary of many is not lemmatised or normalised. In contrast, the aim here is to obtain an embedding model covering the full range of modern English, from the 16th century to present. Additionally, since the goal is to quantify intertextuality, semantic consistency is prioritised: grammatical inflections, which introduce little semantic change, should be reduced to their lemma forms, and historical variant spellings standardised to their modern equivalents.

\begin{figure}[htbp]
	\centering
	\includegraphics[width=\textwidth]{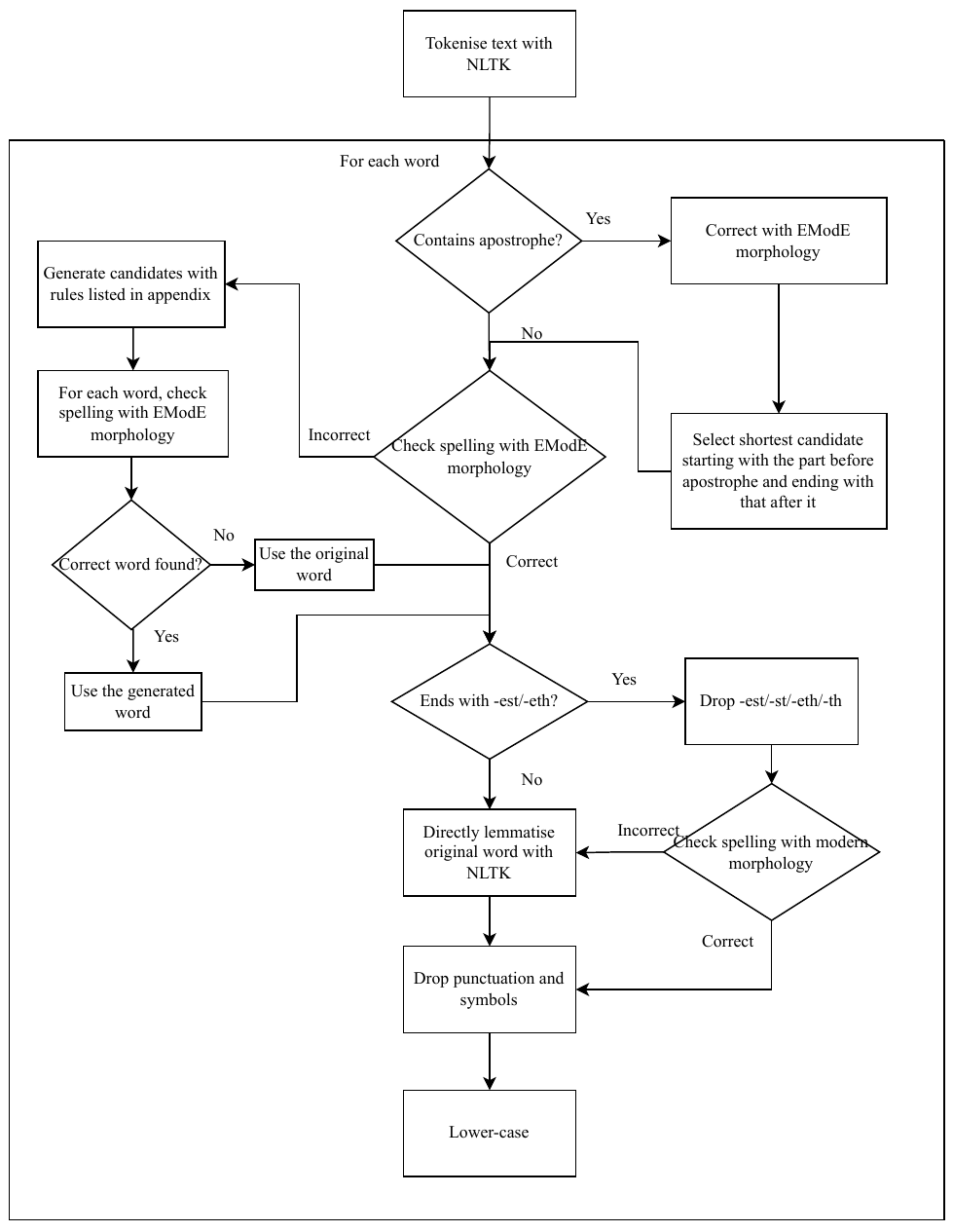}
	\caption{Text cleaning pipeline, composed of three major steps: a) contraction expansion; b) spelling modernisation; c) lemmatisation}
	\label{fig:normalisation}
\end{figure}

Thus, a model was trained on the Oxford Text Archive, which encompasses texts from the entire modern age and includes a diverse array of formats---such as books, pamphlets and tracts---spanning multiple genres, including but not restricted to fiction, religion, philosophy and science. It is acknowledged that lemmatisation and historical text normalisation are by themselves vast, challenging research topics \citep{bollmann_large-scale_2019}. In this work, a simple rule-based approach based on Hunspell \citep{nemeth_hunspell_2014} and NLTK \citep{bird_natural_2009} is employed, which is accurate enough and computationally inexpensive, as shown in Figure~\ref{fig:normalisation}.

After preprocessing the corpus, a Continuous Bag of Words (CBOW) model \citep{mikolov_efficient_2013} was trained using the \textit{Gensim} package \citep{rehurek_software_2010}. A small window size of 3 was opted for to ensure that a word's nearest neighbours would be direct substitutes rather than merely co-occurring within the same topic. Prior work has typically evaluated word embeddings using small synonym selection datasets to assess synonym detection \citep{freitag_new_2005}, or analogy tasks (e.g., Paris - France = Rome - Italy) to test the linearity of the learned vector space \citep{mikolov_efficient_2013}. However, the latter property is not relevant to our task. Furthermore, intertextuality is not limited to synonyms alone---it can arise from near-synonyms, antonyms, and near-antonyms as well. As such, it is more appropriate to evaluate how closely the model aligns with a given thesaurus. While there have been successful attempts at deriving word embedding models from thesauri \citep{jana_can_2018,wang_hg2vec_2022}, to the best of the author's knowledge, this is the first work to use a graph built from thesauri to evaluate word embedding models.

More specifically, an undirected graph is constructed, where each word is represented as a node, and if two words are listed in the same thesaurus entry, they are connected by an edge. Then for each word, find its $k$ nearest neighbours in the vector space of the trained embedding model\footnote{Here $k = 50$, and the similarity metric is cosine.}, and for each neighbour, if it is connected to the original word in the thesaurus graph, it is considered a hit. Given the same thesaurus graph, a better embedding model should yield more hits.

\begin{figure}[htbp]
	\centering
	\begin{tikzpicture}
		\begin{axis}[
			width=0.8\textwidth,
			height=0.6\textwidth,
			xlabel={Embedding Dimension},
			ylabel={\# Hits},
			grid=both,
			grid style={line width=.1pt, draw=gray!10},
			major grid style={line width=.2pt,draw=gray!50},
			thick,
			mark size=2pt
		]
		\addplot[
			mark=*,
			]
			coordinates {
				(150, 319085)
				(250, 378901)
				(320, 387057)
				(330, 387804)
				(340, 389688)
				(350, 390917)
				(360, 391561)
				(370, 392364)
				(380, 392681)
				(450, 394515)
			};
		\end{axis}
	\end{tikzpicture}
	\caption{Relation between the embedding dimension and the thesaurus graph hit score}
	\label{fig:dimension}
\end{figure}
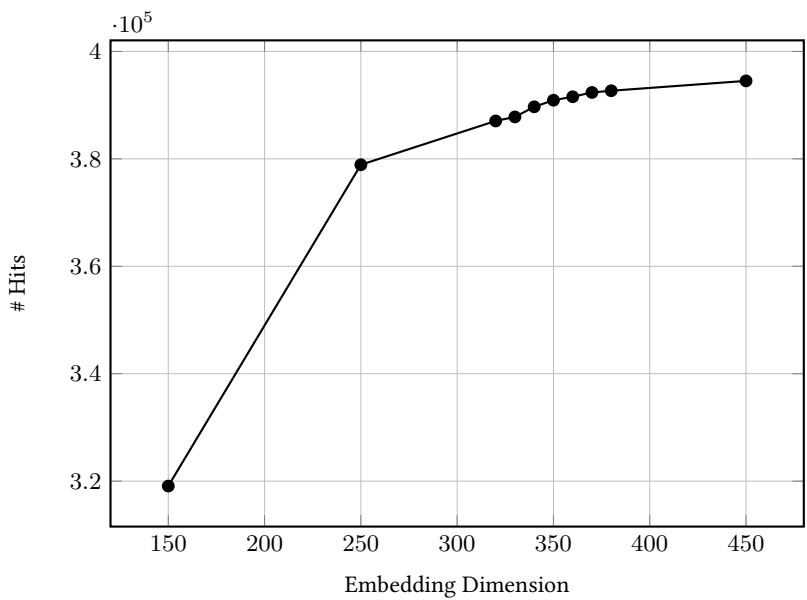

The Merriam-Webster Unabridged Thesaurus \citep{MW_Thesaurus} was used and it was found that the performance of the model increases with the vector space's dimensionality, as demonstrated in Figure~\ref{fig:dimension}, although a larger dimensionality will increase subsequent computational cost. Therefore, a careful balance should be struck between performance and efficiency, and a dimensionality of $350$ was chosen. Manual inspection on certain words reveals that the model indeed captures all major historical usage. For example, the older senses of `behaviour,' `deportment,' and `intercourse,' `acquaintance,' `intimacy' together with the current sense of `communication,' `discourse' all appear in the 20-nearest neighbours of the word `conversation.'

\section{Validation study}

In order to verify this model of intertextuality, four books with differing degrees of intertextuality---George Herbert's \textit{Temple}, John Donne's complete poems, and Charlotte Brontë's \textit{Professor} and \textit{Villette}---were selected. The first two are quintessential metaphysical poetry, with the former centred on religious themes, and the latter covering diverse topics including love, faith, politics and contemporary society. The two of Brontë's novels share the same diction, style and materials, as well as the common themes of independence, self-help and romance. They also contain occasional outpourings of religious sentiments and abundant biblical references. Thus, the four texts make an interesting validation study.

The texts were first processed with the same pipeline that was applied to the training corpus, and the 500 most frequent words in the corpus were used as the stop-list: if more than half of the words in an n-gram were in the list, then it was dropped. As described previously, there were two hyperparameters, namely the size of n-grams and the similarity threshold. It was found that their choices do not have much influence on the results qualitatively, as shown in the heat maps in Table~\ref{tab:validation}, and the results align with established literary interpretation: the two novels by the same author, Brontë, which share characters and themes, show the highest degree of intertextuality; the two metaphysical poetry collections, while stylistically related, are less connected, and the cross-genre, cross-period comparisons are lowest. Moreover, it is worth mentioning that too high a threshold (e.g., 0.5) will make the difference between the intertextuality scores less pronounced.

\begin{table}[t]
	\centering
	\caption{The experimental results under various hyperparameters all indicate that \textit{The Professor} and \textit{Villette} are the most intertextually related to each other, followed by Herbert's and Donne's poetry; Donne's works have a weaker link with those of Brontë; the intertextuality between Herbert's \textit{Temple} and Brontë's novels is the weakest.}
	\begin{tabular*}{\textwidth}{ccccc}
		
		\hline

		\diagbox{\textbf{Threshold}}{$n$} & 3 & 4 & 5 & 6 \\ \hline
		
		0.01 & \includegraphics[width=0.16\textwidth]{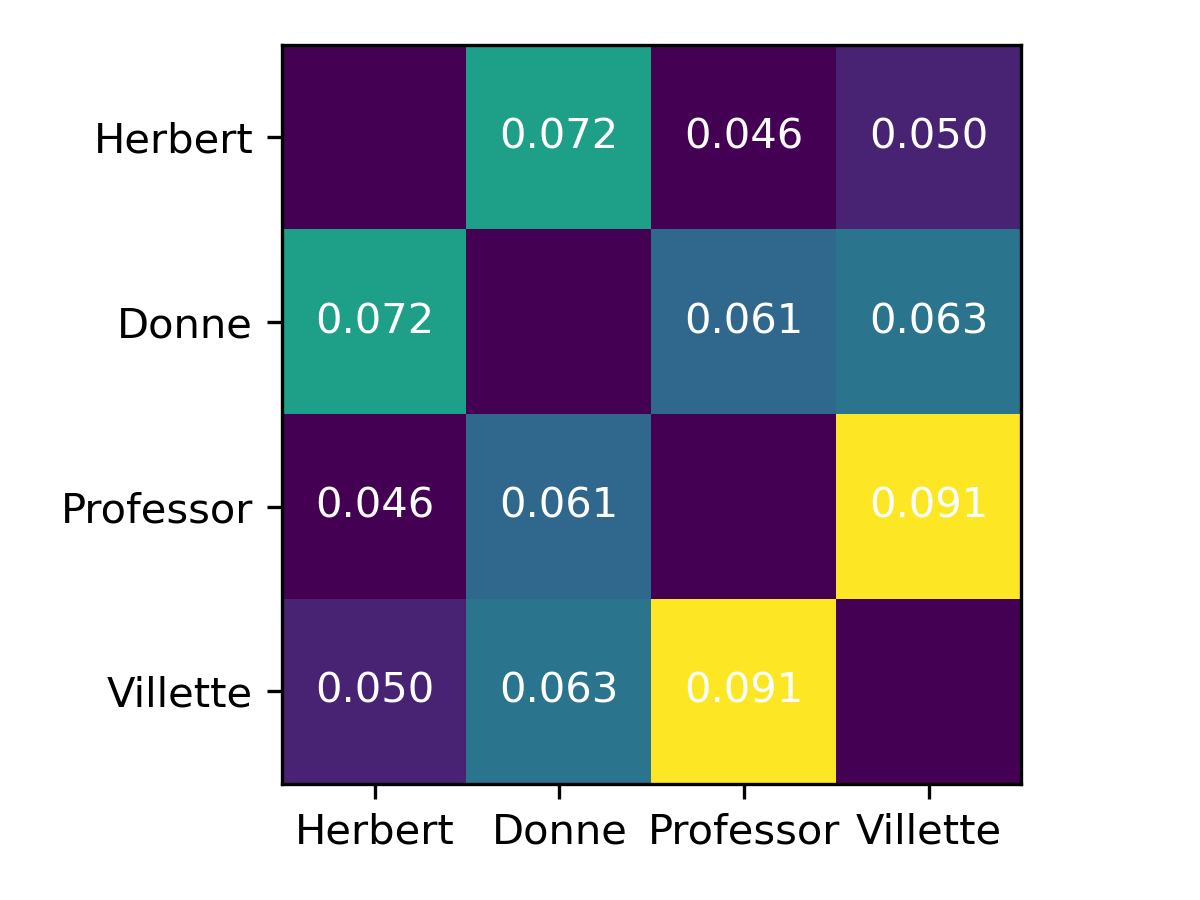} & \includegraphics[width=0.16\textwidth]{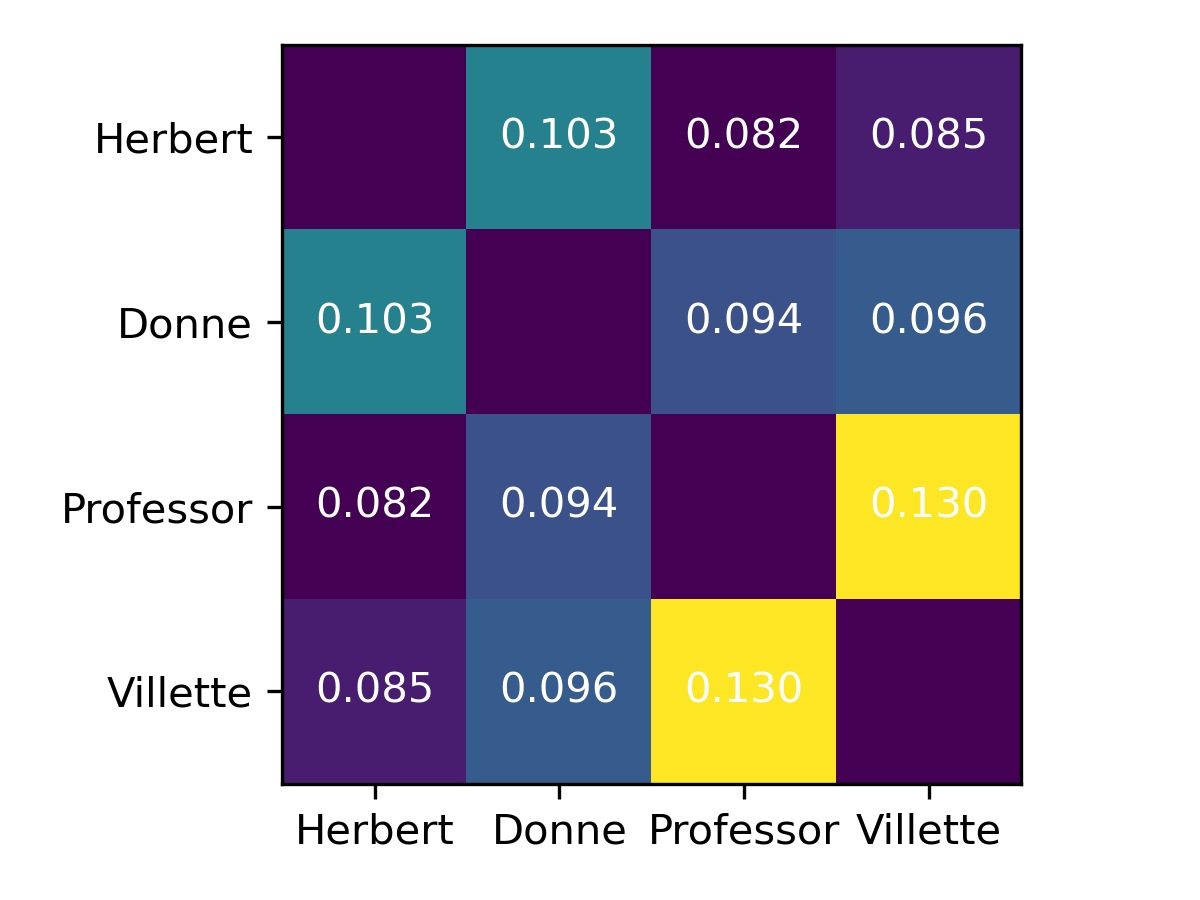}  & \includegraphics[width=0.16\textwidth]{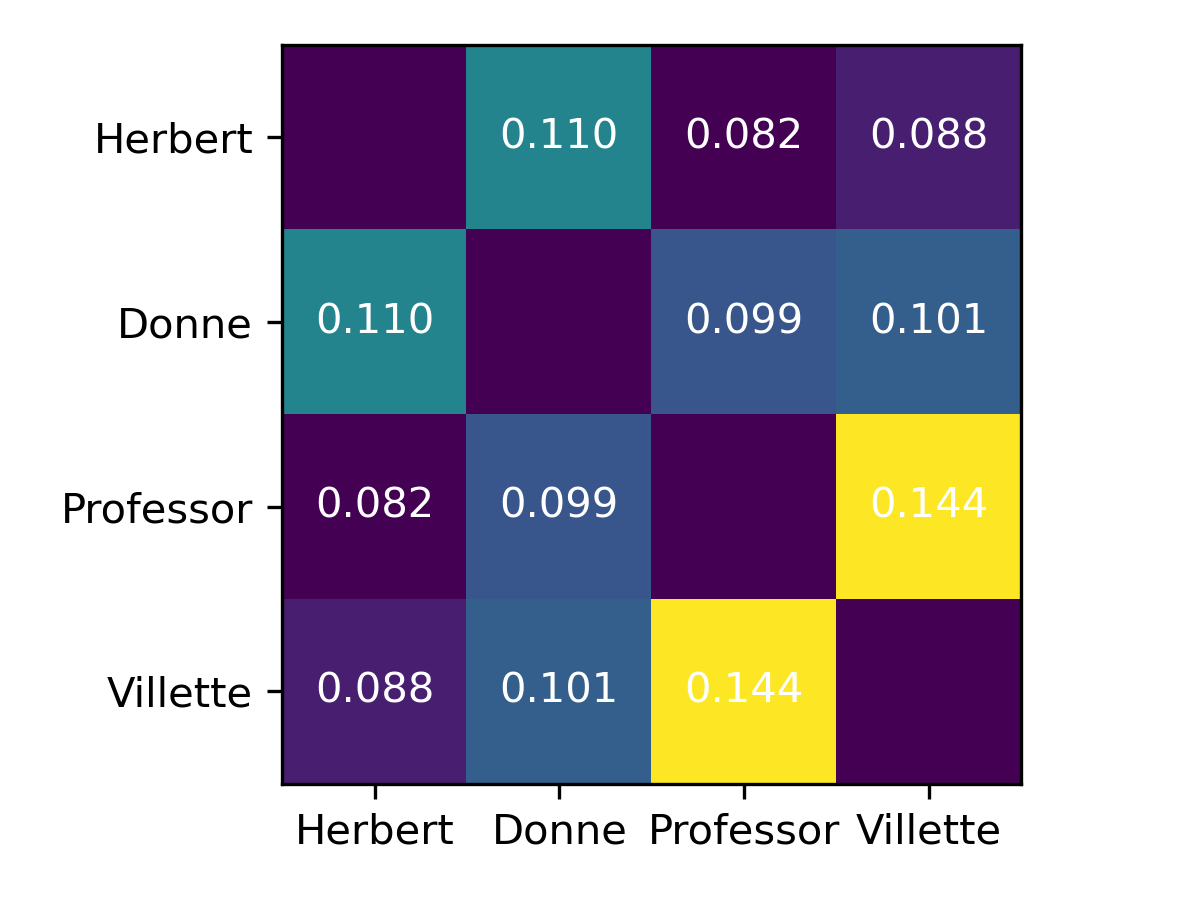}  &  \includegraphics[width=0.16\textwidth]{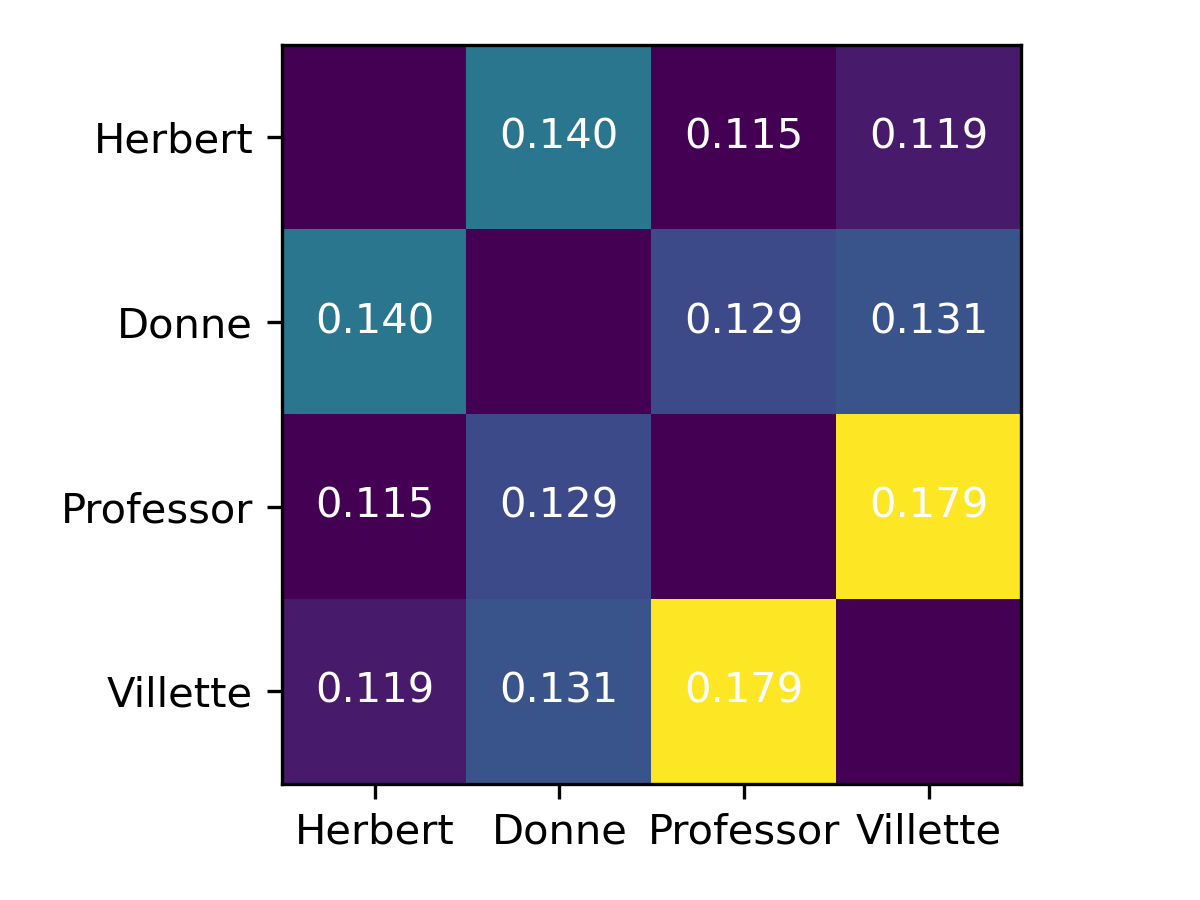} \\

		0.05 & \includegraphics[width=0.16\textwidth]{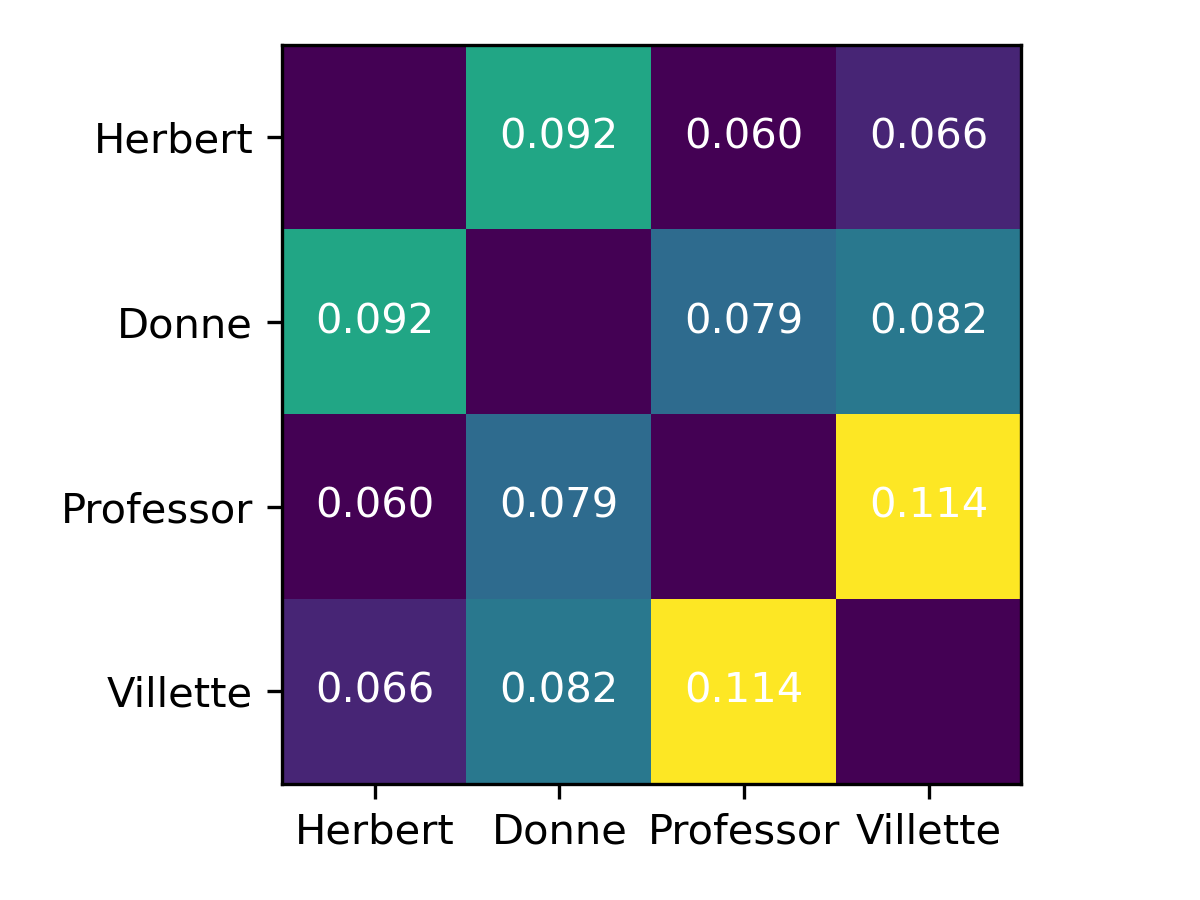}  &  \includegraphics[width=0.16\textwidth]{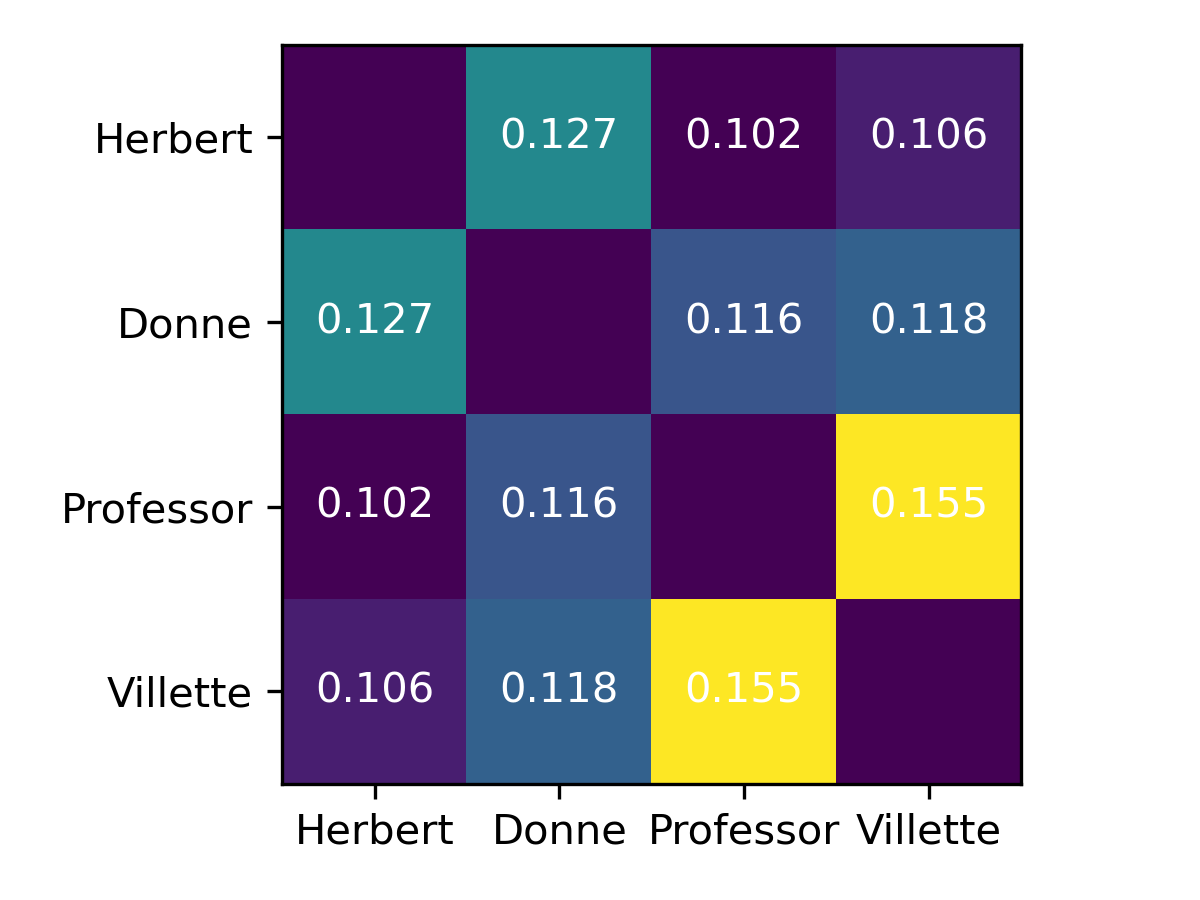} & \includegraphics[width=0.16\textwidth]{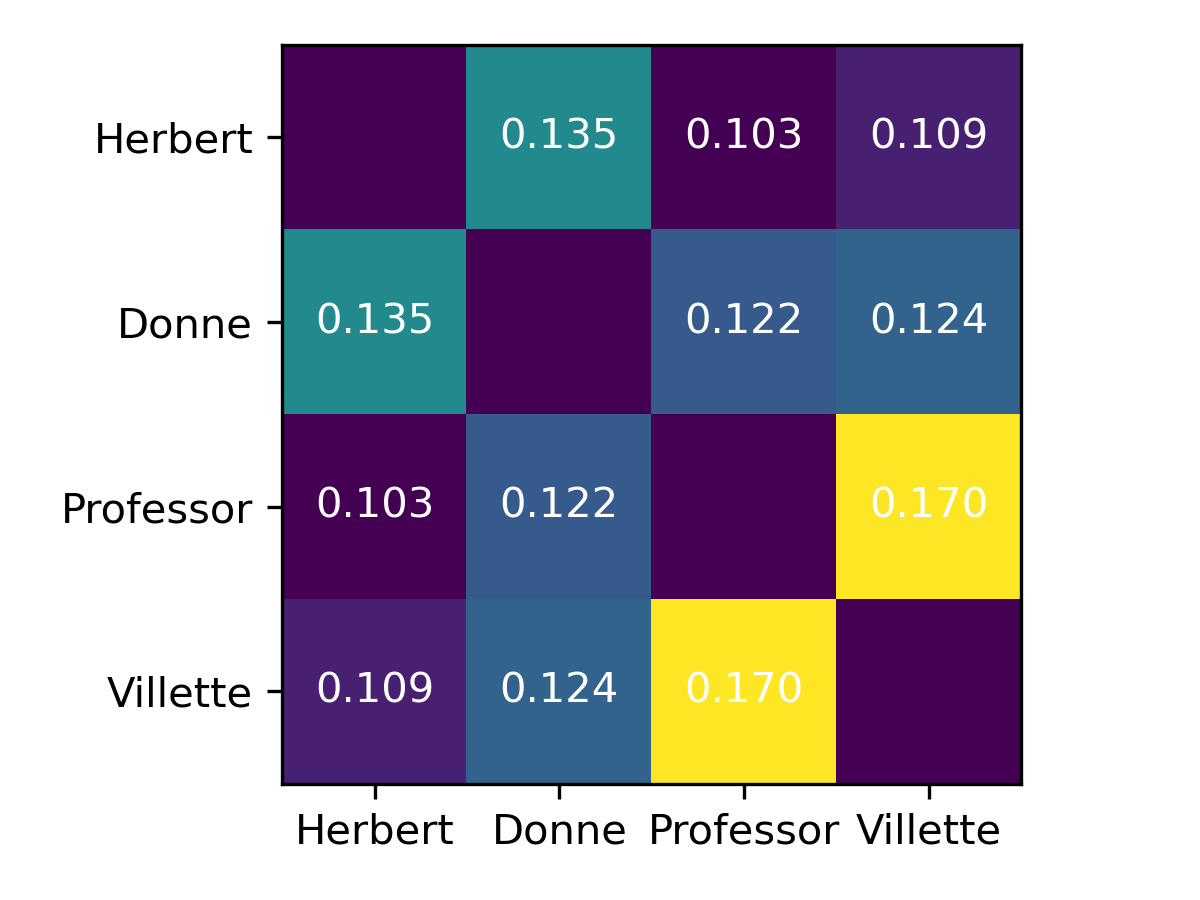}  &  \includegraphics[width=0.16\textwidth]{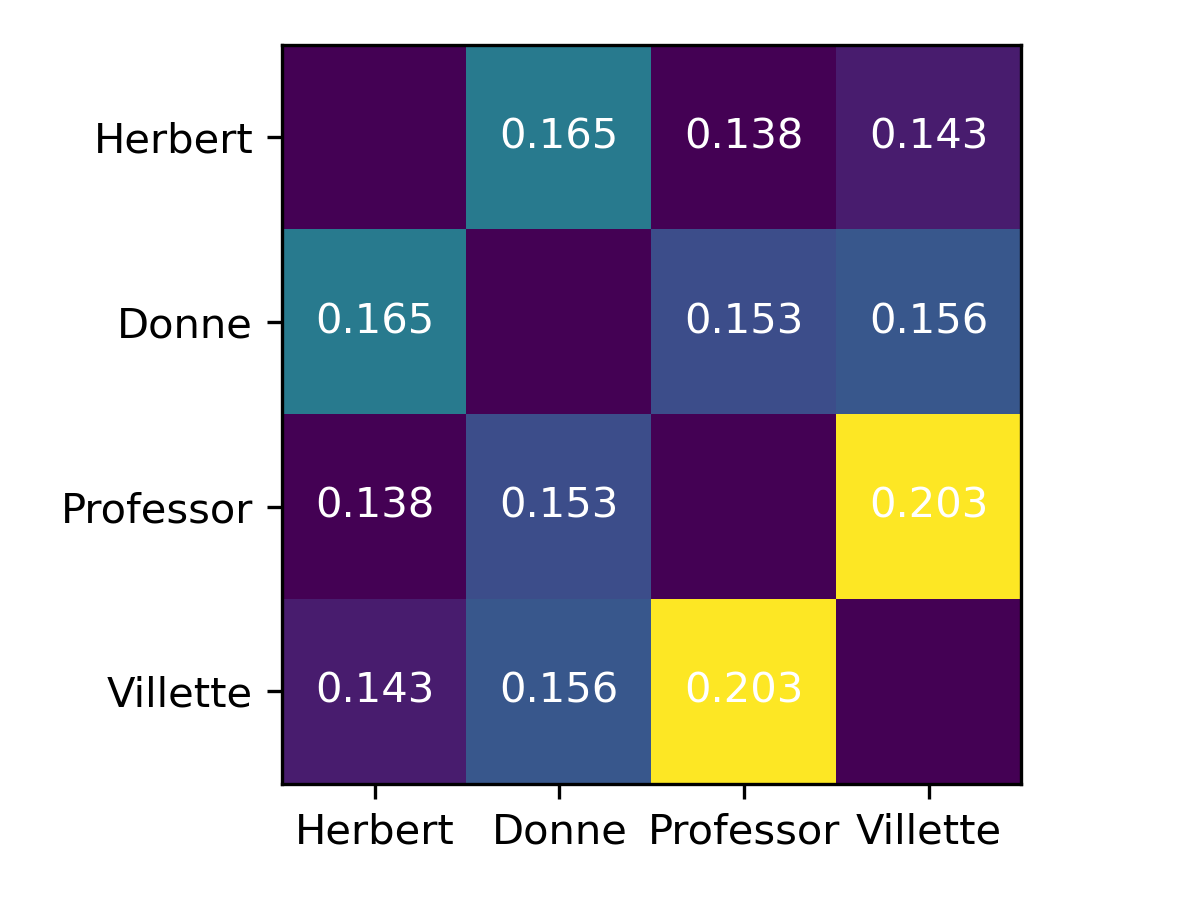} \\ 

		0.10  &  \includegraphics[width=0.16\textwidth]{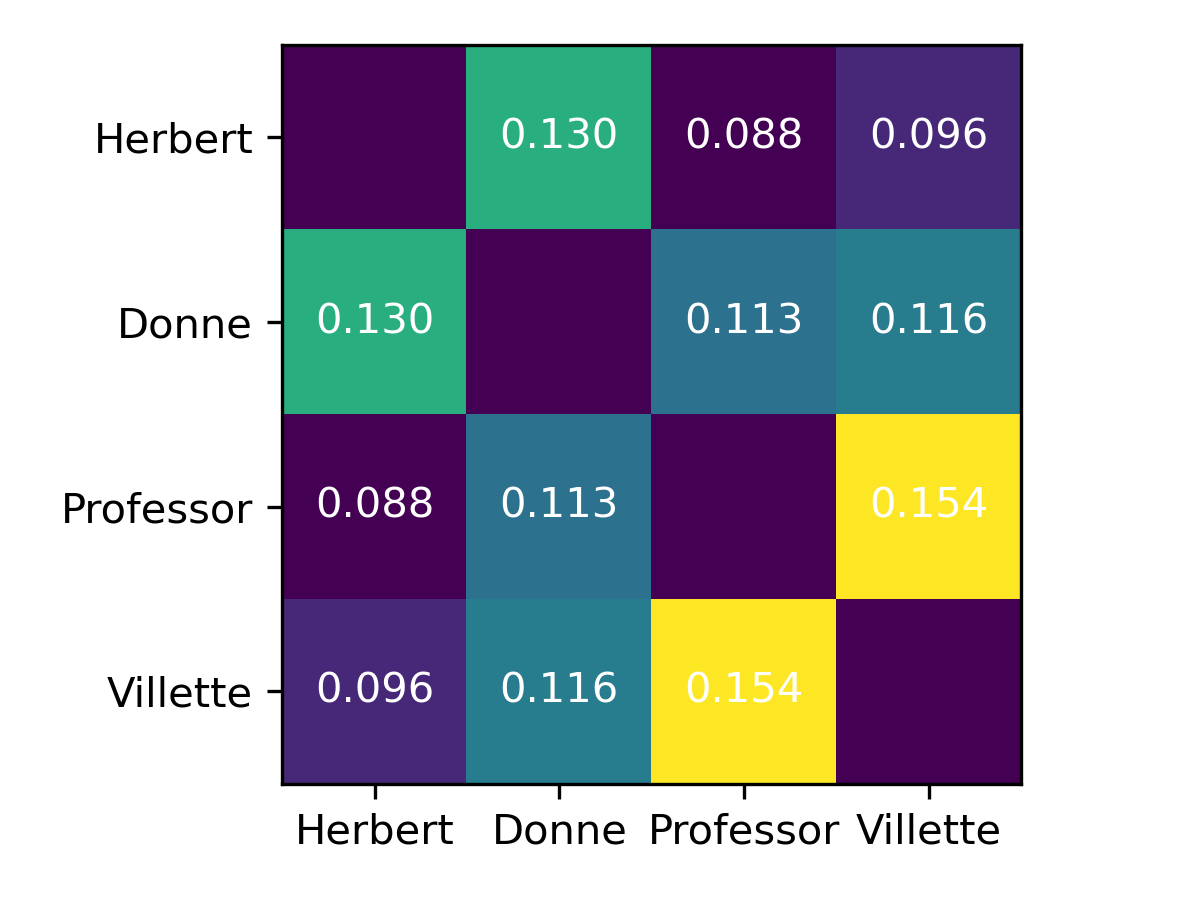} &  \includegraphics[width=0.16\textwidth]{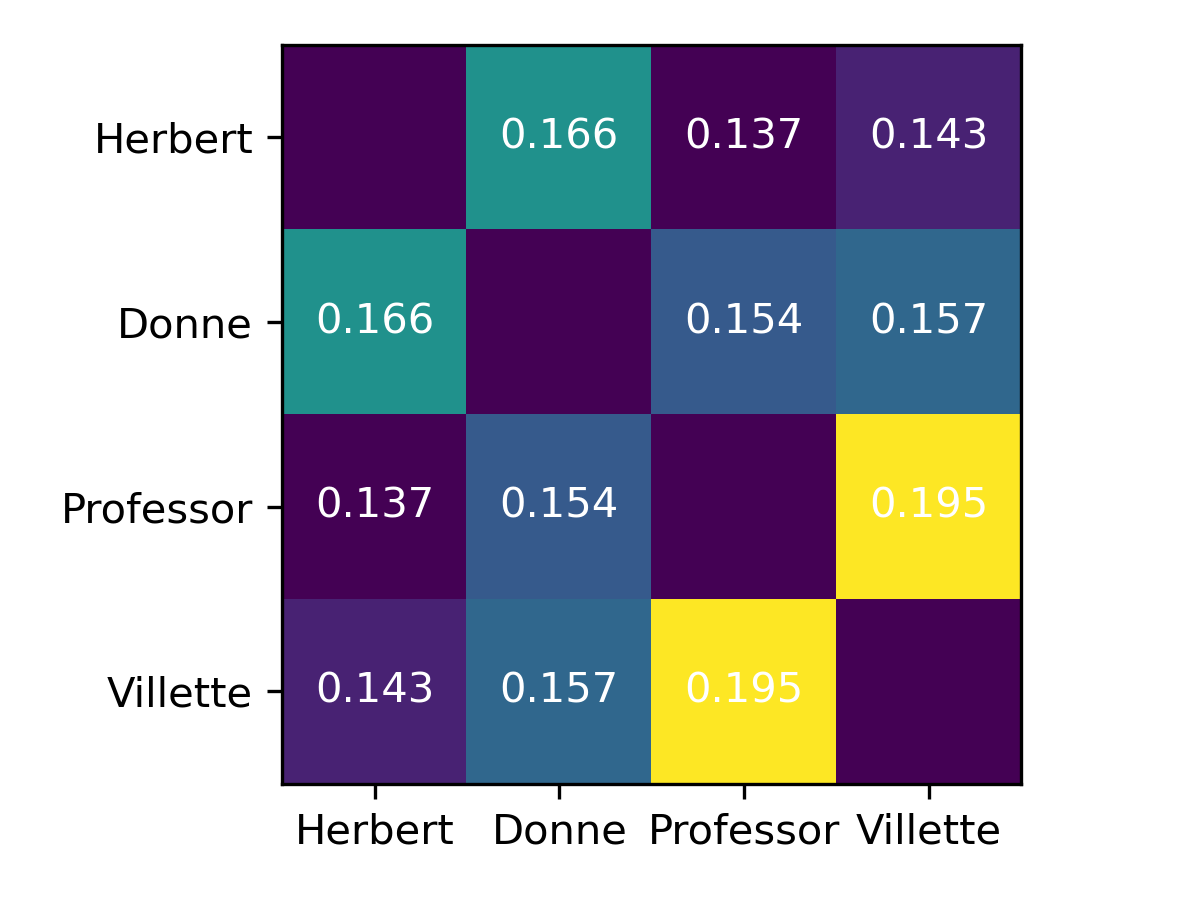} & \includegraphics[width=0.16\textwidth]{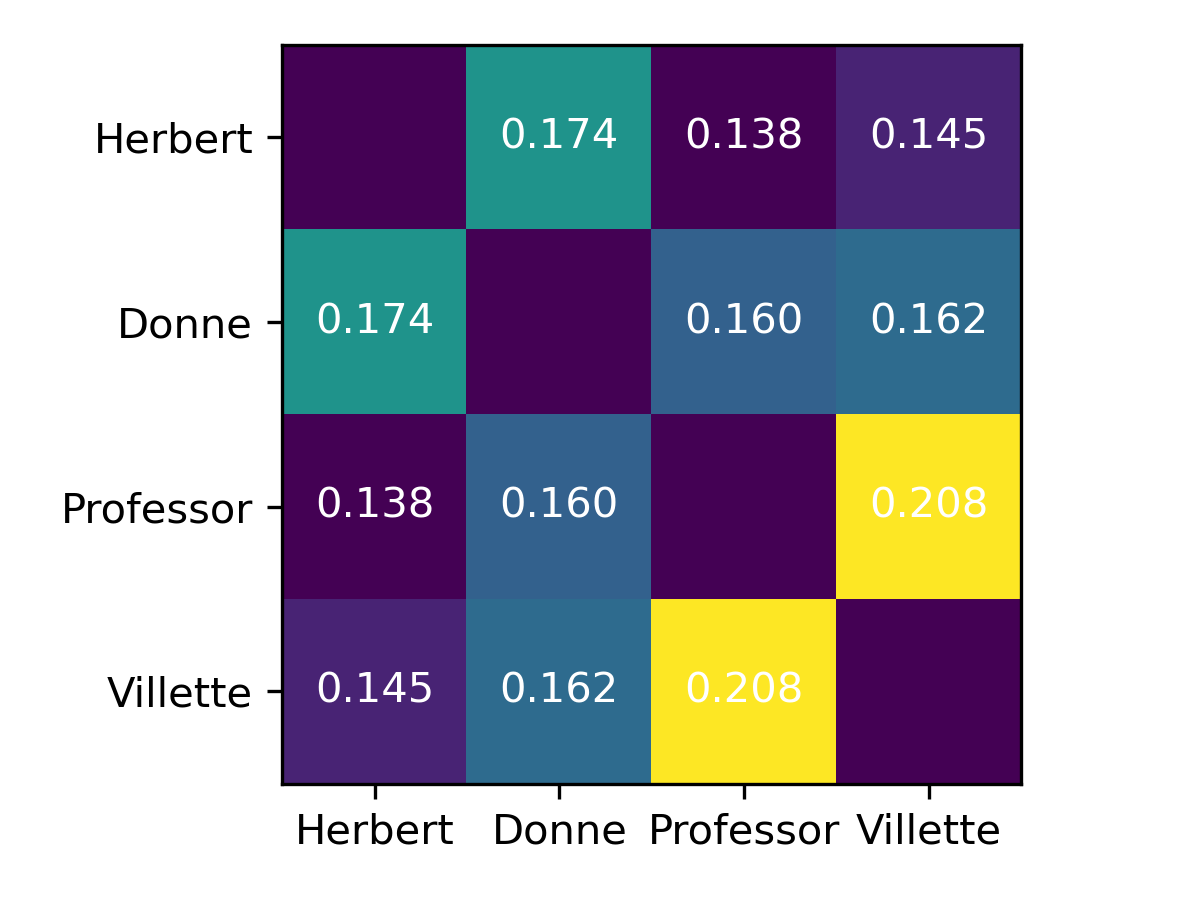}  & \includegraphics[width=0.16\textwidth]{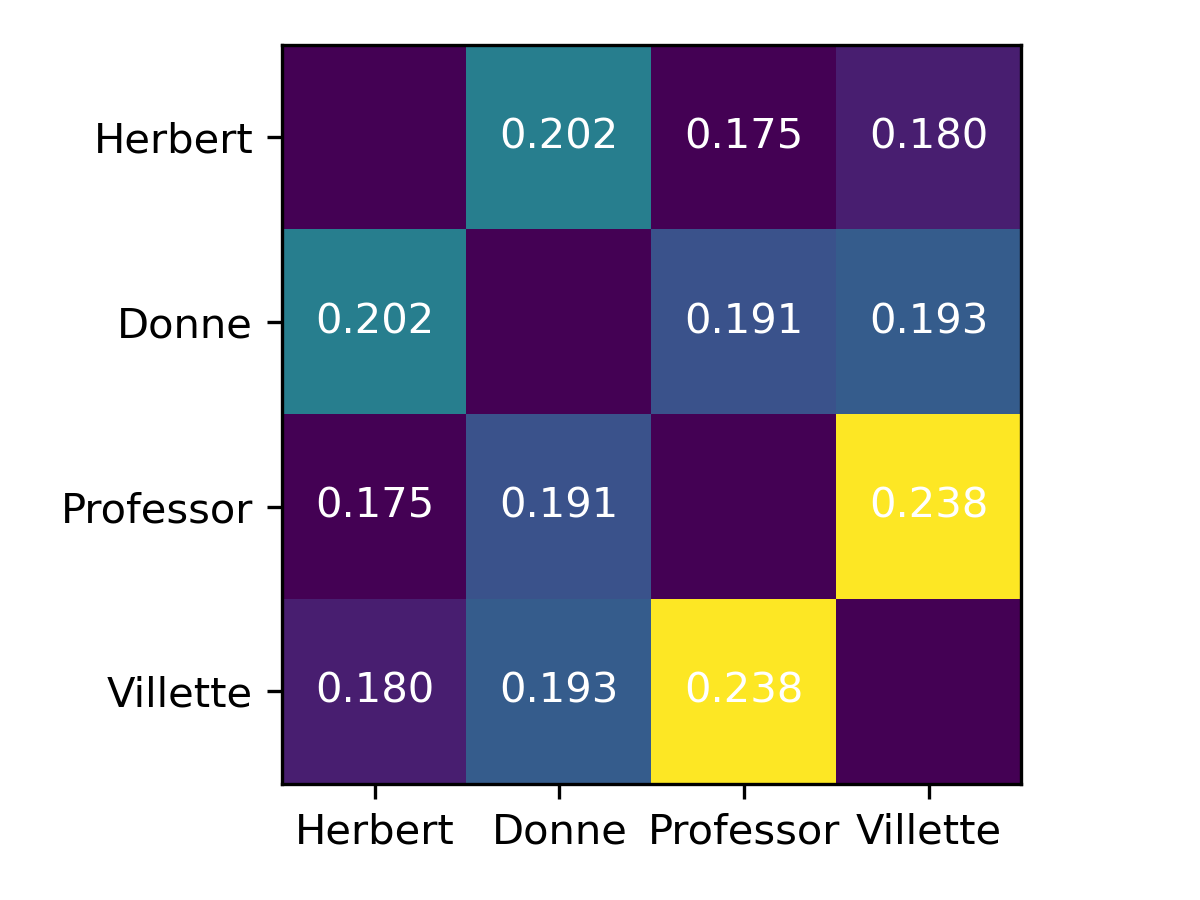}  \\

		0.20  &  \includegraphics[width=0.16\textwidth]{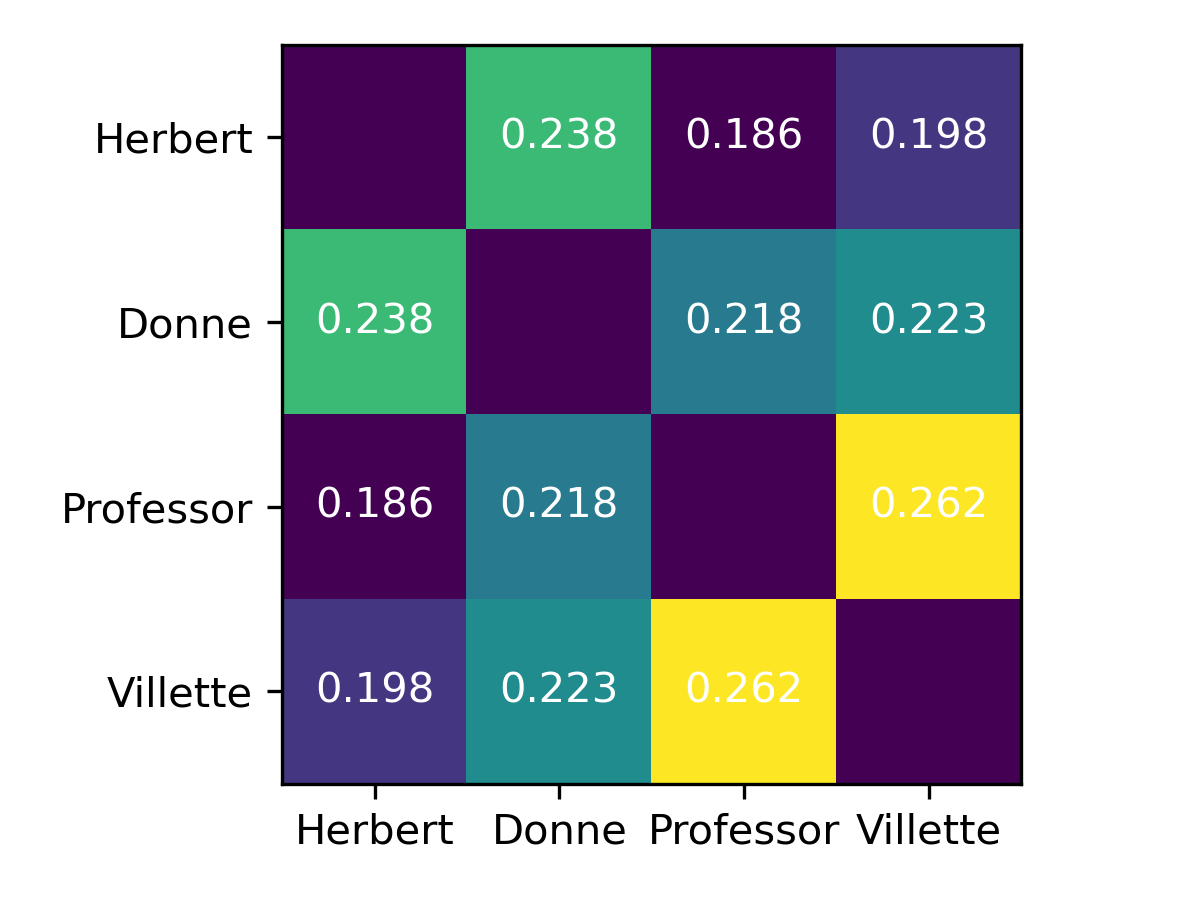} & \includegraphics[width=0.16\textwidth]{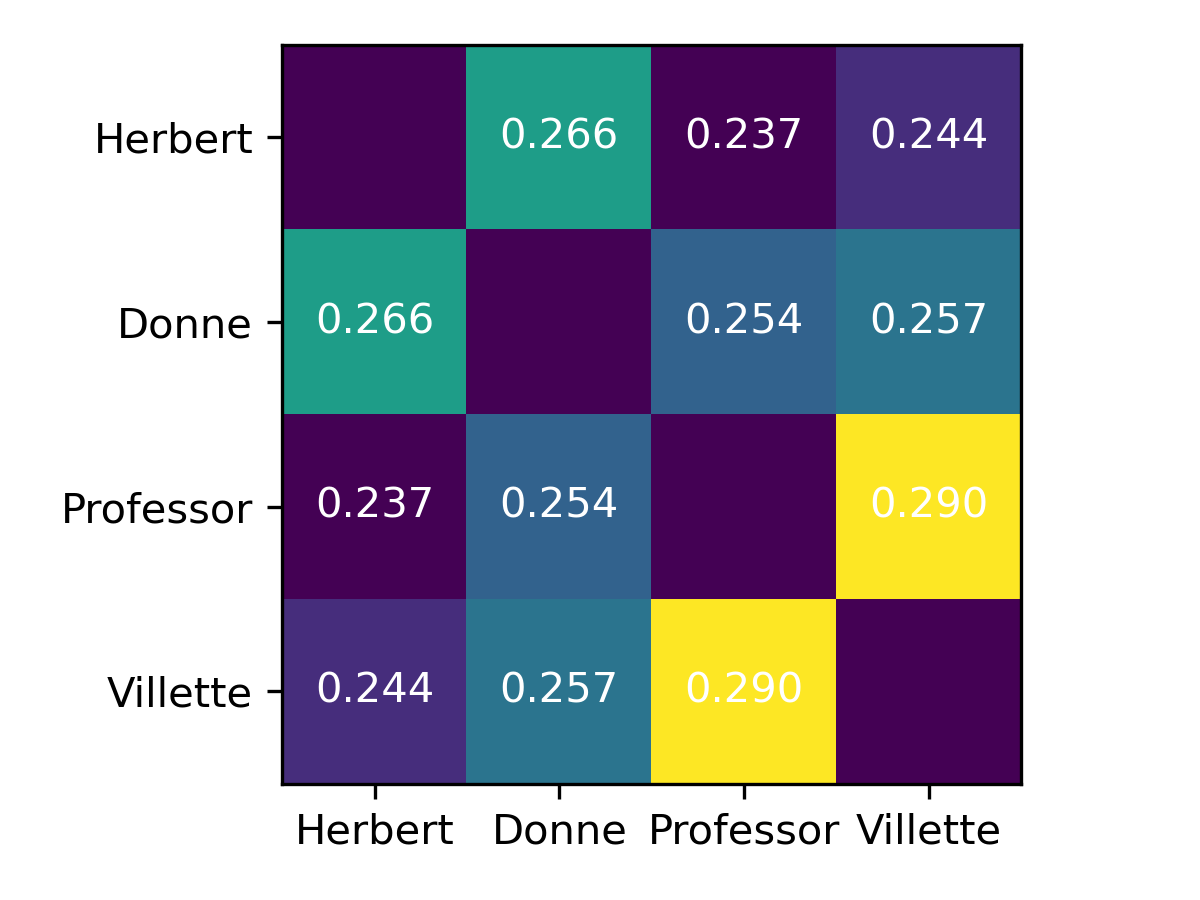}  & \includegraphics[width=0.16\textwidth]{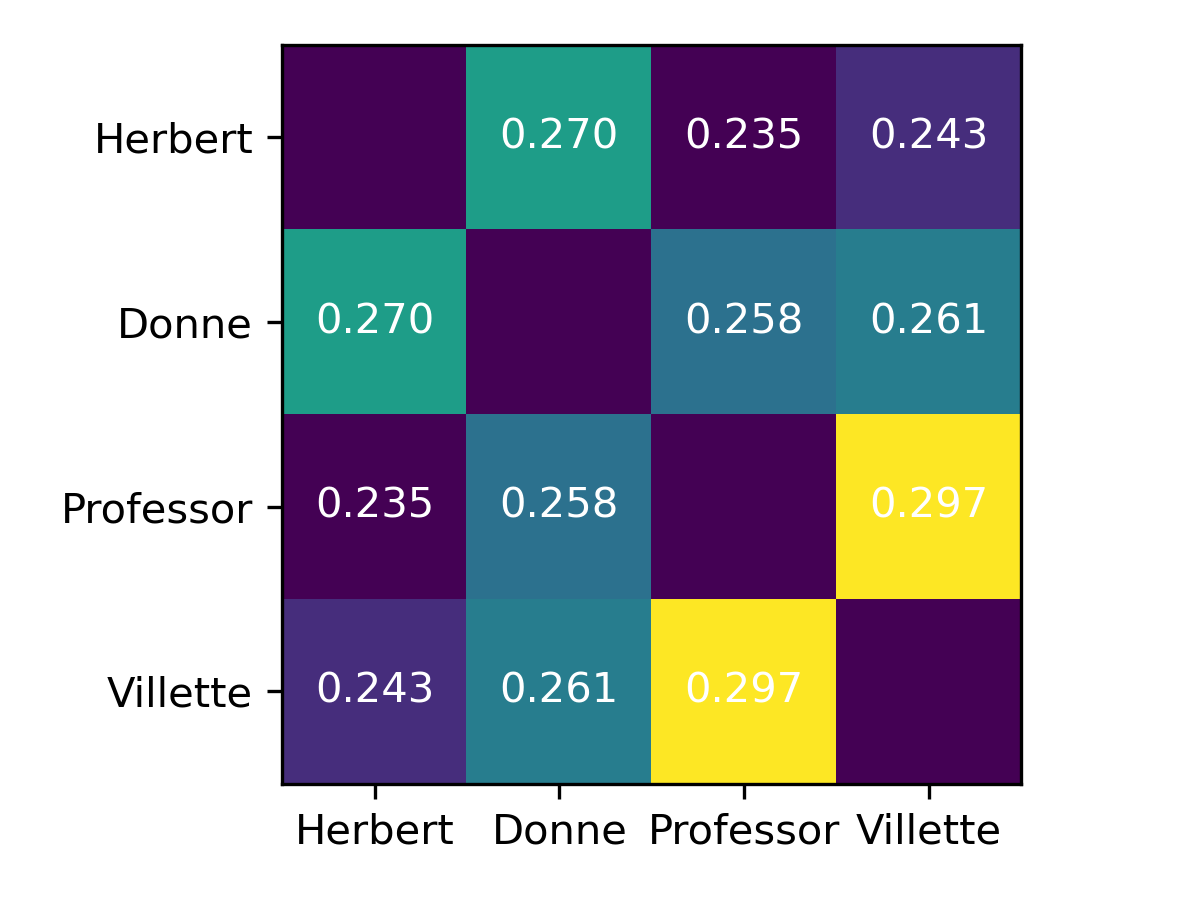}  &  \includegraphics[width=0.16\textwidth]{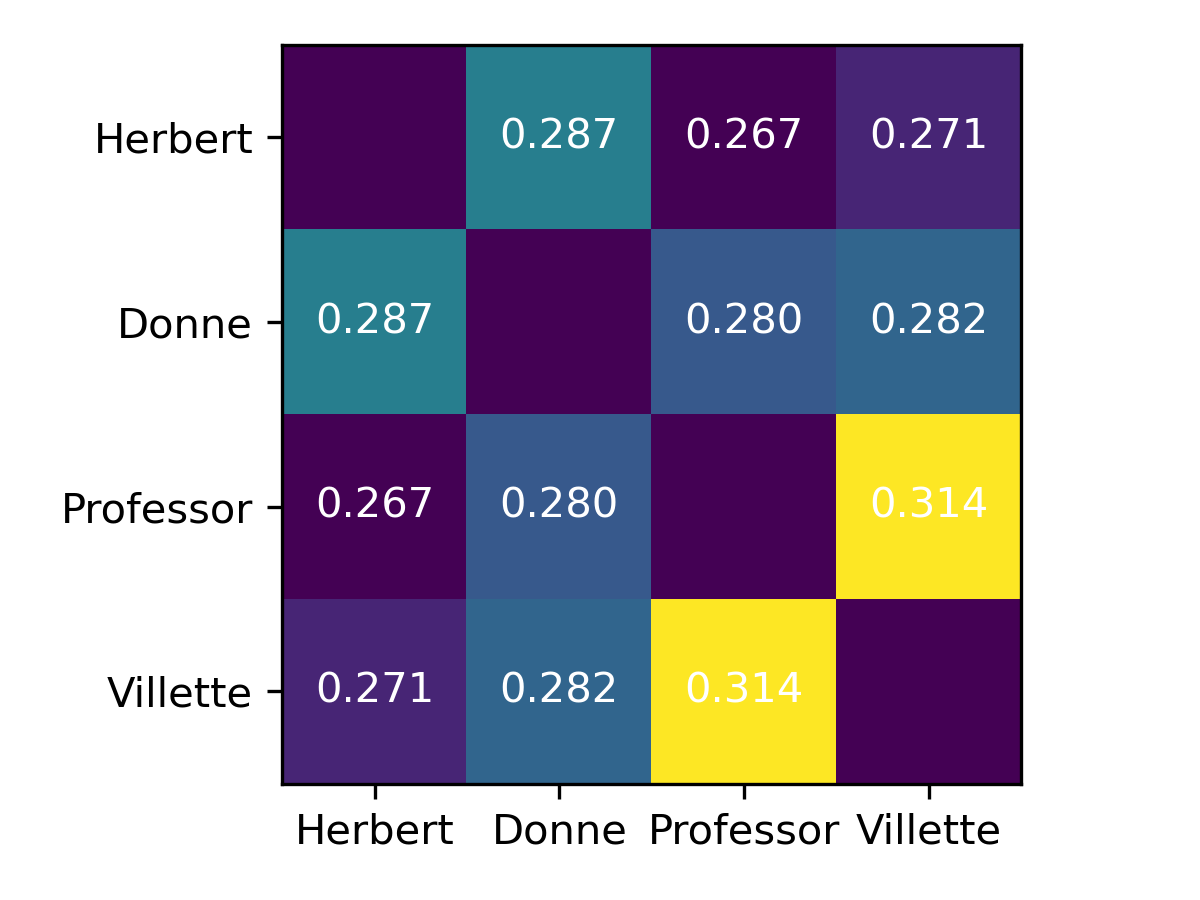} \\ 

		0.50  & \includegraphics[width=0.16\textwidth]{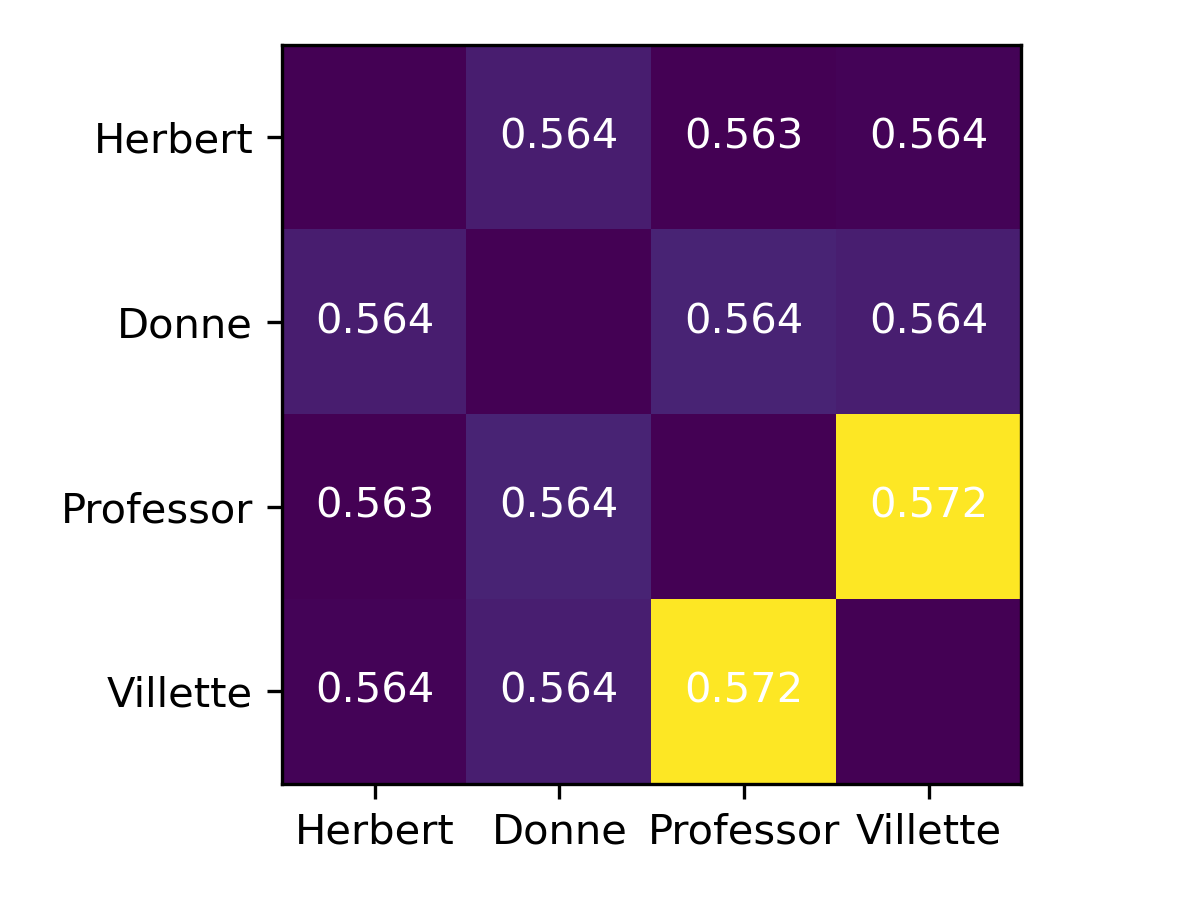}  &  \includegraphics[width=0.16\textwidth]{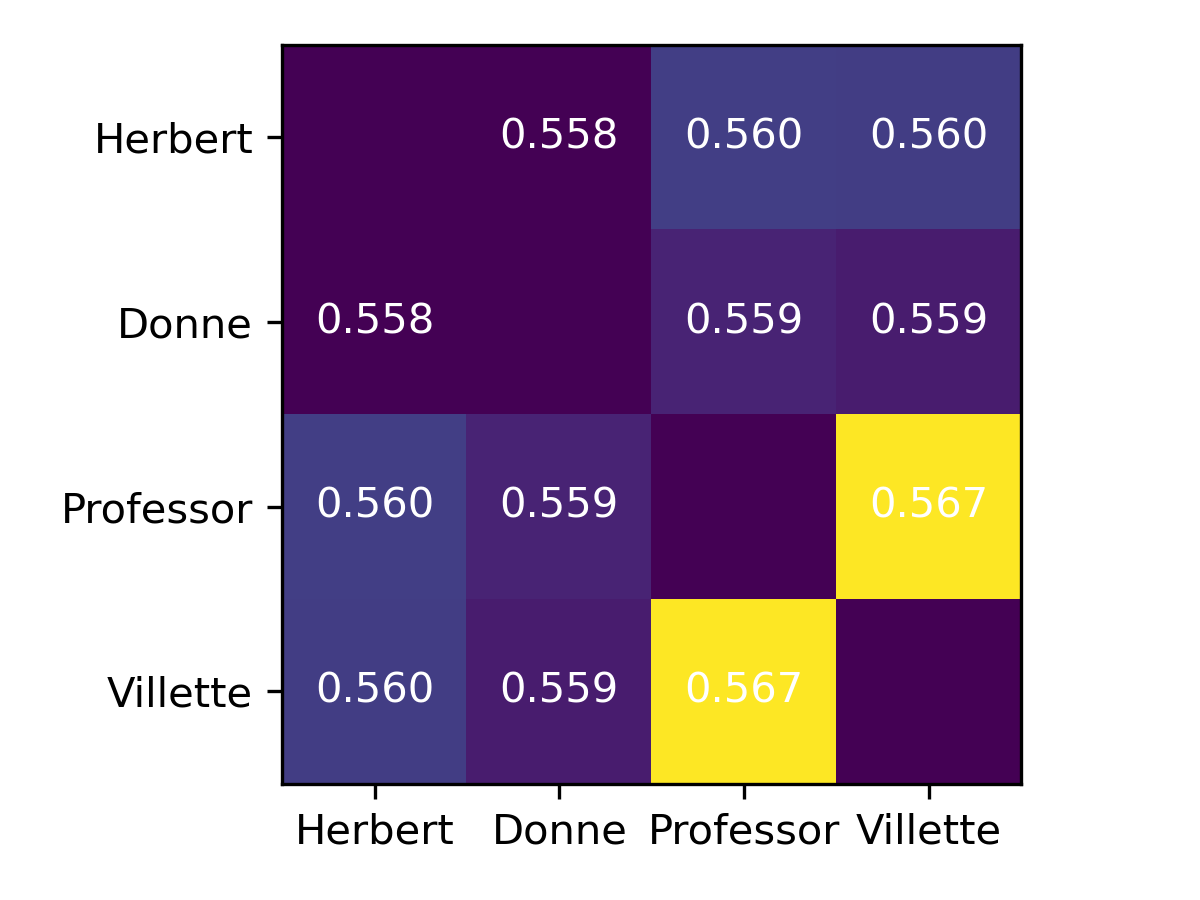} &  \includegraphics[width=0.16\textwidth]{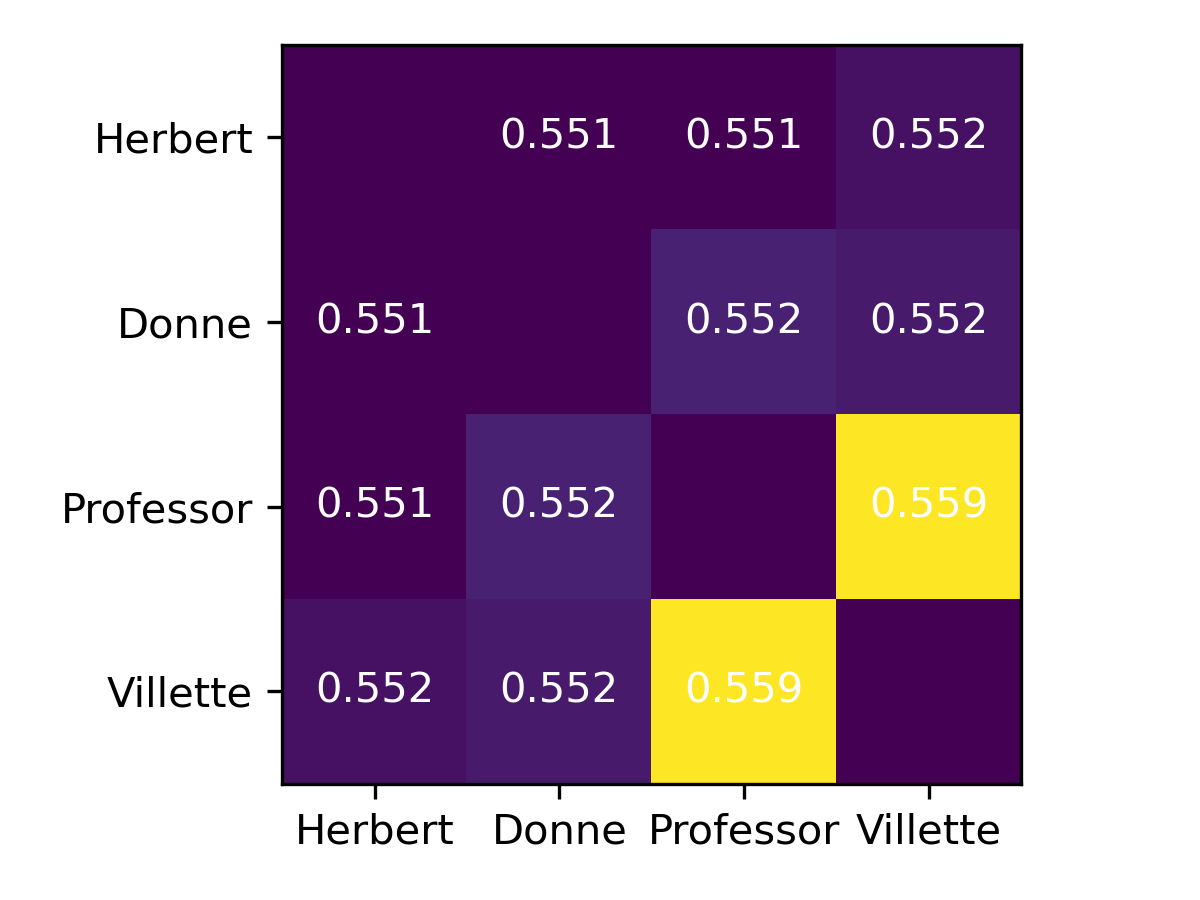} & \includegraphics[width=0.16\textwidth]{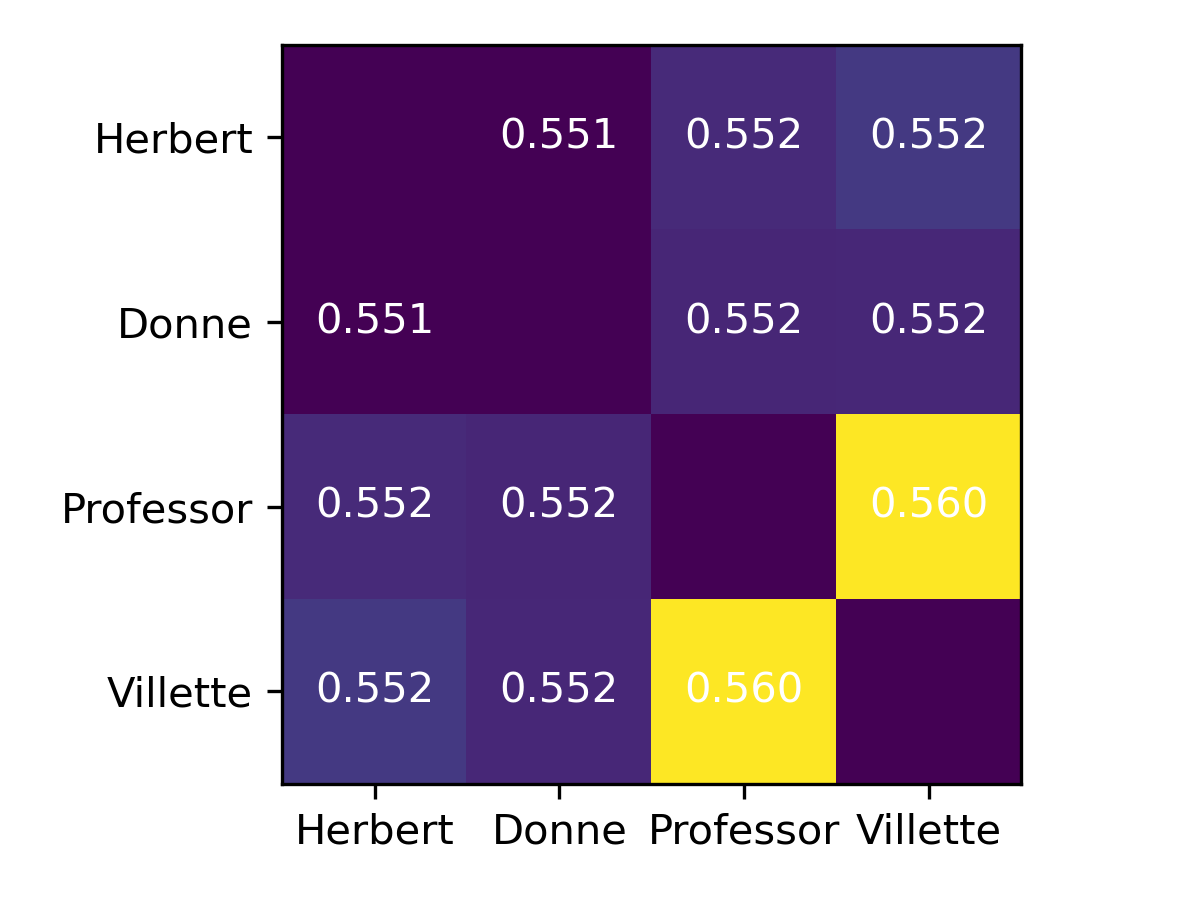}  \\ \hline
	\end{tabular*}
	\label{tab:validation}
\end{table}

\section{Scalability}

Scalability is a focal point of previous studies. Accordingly, to evaluate the performance of this method at scale, 267 works of different lengths, time periods and genres were selected, and the pairwise intertextuality scores were computed with $n$ and the threshold set to 3 and 0.2, respectively. The entire computation, involving 35,511 comparisons, was completed on a single workstation with one NVIDIA GeForce RTX 4060 Ti GPU in 49 hours.

The number of document-level comparisons grows quadratically with the number of documents in a given corpus, and each comparison's runtime scales with the product of the numbers of n-grams in the two documents. Despite this, the method remains tractable due to its high degree of parallelisability. With appropriate batching and vectorisation, speedups of up to 5,000$\times$ were observed, compared to a naïve nested-loop implementation.

\begin{algorithm}
    \begin{algorithmic}[1]
        \State \textbf{Inputs} Texts $A = (\mathrm{a}_1, \mathrm{a}_2, \mathrm{a}_3, \ldots, \mathrm{a}_{m_A})$, $B = (\mathrm{b}_1, \mathrm{b}_2, \mathrm{b}_3, \ldots, \mathrm{b}_{m_B})$, similarity threshold $\tau$
        \State \textbf{Output} $\mathrm{Intertextuality}(A, B)$
        \State $\mathrm{A} \gets [\mathrm{a}_1; \mathrm{a}_2; \ldots; \mathrm{a}_{m_A}]^\mathrm{T}$
        \State $\mathrm{B} \gets [\mathrm{b}_1; \mathrm{b}_2; \ldots; \mathrm{b}_{m_B}]^\mathrm{T}$
        \State $\mathrm{P} \gets \mathrm{A}\mathrm{B}^\mathrm{T}$
        \State $\mathrm{n}_\mathrm{A} \gets |\mathrm{A}|$
        \State $\mathrm{n}_\mathrm{B} \gets |\mathrm{B}|$
        \State $\mathrm{Q} \gets \mathrm{n}_\mathrm{A} \otimes \mathrm{n}_\mathrm{B}$
        \State $\mathrm{cosine}(\mathrm{A}, \mathrm{B}) \gets \mathrm{P} \oslash \mathrm{Q}$
        \State $\mathrm{T} \gets |\mathrm{cosine}(\mathrm{A}, \mathrm{B})| > \tau$
        \State $\mathrm{S} \gets \mathrm{cosine}(\mathrm{A}, \mathrm{B}) \cdot \mathrm{T}$
        \State $S \gets \sum_{i=1}^{m_A} \sum_{j=1}^{m_B} \mathrm{S}_{ij}$
        \State $c \gets \sum_{i=1}^{m_A} \sum_{j=1}^{m_B} \mathrm{T}_{ij}$
        \If{$c > 0$}
            \State $\mathrm{Intertextuality}(A, B) \gets \frac{S}{c}$
        \Else
            \State $\mathrm{Intertextuality}(A, B) \gets -1$
        \EndIf
        \State \Return $\mathrm{Intertextuality}(A, B)$
    \end{algorithmic}
    \caption{A vectorised implementation of the algorithm, where $\mathrm{u} \otimes \mathrm{v}$ is the outer product of two vectors, and $\mathrm{A} \oslash \mathrm{B}$ is the element-wise division of two matrices}
    \label{alg:vectorised}
\end{algorithm}

As shown in Algorithm~\ref{alg:vectorised}, at the core of this efficiency is the use of matrix operations to compute pairwise cosine similarities. For each document, the n-gram embeddings are stacked into an $m \times d$ matrix, where $m$ is the number of n-grams and $d$ the embedding dimension, and the $\mathrm{L}_2$ norms of its rows are first computed. To compare two documents, we compute the product of their respective n-gram matrices, then divide the result element-wise by the outer product of their norms. The result is a dense similarity matrix containing all pairwise n-gram similarities between the two documents.

For very large documents whose n-gram vectors cannot fit into memory, we divide them into sub-documents. Intertextuality scores are then computed for each sub-pair and averaged. And the process can be easily distributed across multiple machines since the computation order is irrelevant.

Memory consumption can be further reduced by using 32-bit floating-point precision, which was found to be both efficient and accurate. In contrast, 16-bit precision, while theoretically appealing for its even smaller footprint, significantly slowed down computation on our test hardware.

\section{Network analysis}

Having computed the intertextuality between all document pairs, we can derive a complete weighted graph from it, where the nodes are the documents and the edges' weights are the intertextuality scores. The graph can be directed with later works pointing at earlier works, when it can be interpreted as a network of literary influence \citep{fedchenko_a_2024}, or undirected, when it more closely resembles the reader's notion of intertextuality---a chronologically earlier work can also call a later one to mind. In this paper, centrality analysis was performed on the directed version of the network and community detection on the undirected one. The NetworkX implementation of the algorithms \citep{hagberg_exploring_2008} was used. The full results are published in the accompanying GitHub repository (see Appendix~\ref{sec:data}).

For centrality analysis, five measures were used: eigenvector, PageRank, Laplacian, closeness and harmonic centrality. The results of the algorithms (including under different damping factors for PageRank and Laplacian centrality) were largely identical, and it was found that the centrality is mostly related to a text's year of publication, which is plausible, since earlier texts can be referenced by all later ones. However, more influential ones like Goethe's \textit{Sorrows of Young Werther} were indeed ranked higher among contemporaries, as shown in Table~\ref{tab:pagerank}.

\begin{table}[h]
	\centering
	\caption{Undamped PageRank of the texts. Earlier texts were typically ranked higher, and influential ones were higher than their contemporaries.}
	\begin{tabular*}{\textwidth}{cc}
	\hline
	\multicolumn{1}{c}{\textbf{Text}}                                                               &\textbf{PageRank} \\ \hline
	\multicolumn{1}{c}{Dante Alighieri -- The Divine Comedy   (1320)}                       & 0.1605   \\ 
	\multicolumn{1}{c}{Petrarch -- Canzoniere (1353)}                                       & 0.0784   \\ 
	\multicolumn{1}{c}{Giovanni Boccaccio -- The Decameron (1358)}                          & 0.0571   \\ 
	\multicolumn{2}{c}{…}                                                                             \\ 
	\multicolumn{1}{c}{Jonathan Edwards -- The Freedom of the   Will (1754)}                & 0.0022   \\
	\multicolumn{1}{c}{Jean-Jacques Rousseau -- Julie, or the New  Héloïse (1761)}         & 0.0022   \\ 
	\multicolumn{1}{c}{Jean-Jacques Rousseau -- Émile (1762)}                               & 0.0022   \\ 
	\multicolumn{1}{c}{Johann Wolfgang von Goethe -- The Sorrows of Young Werther (1774)} & 0.0021   \\ 
	\multicolumn{1}{c}{Voltaire -- Candide (1759)}                                          & 0.0021   \\ 
	\multicolumn{1}{c}{Charles Montesquieu -- The Spirit of the   Laws (1748)}              & 0.0021   \\ 
	\multicolumn{1}{c}{Jean-Jacques Rousseau -- The Social   Contract (1762)}               & 0.0020    \\
	\multicolumn{2}{c}{…}                                                                             \\ 
	\multicolumn{1}{c}{T. S. Eliot -- Four Quartets (1943)}                                 & 0.0006   \\ 
	\multicolumn{1}{c}{Evelyn Waugh -- Brideshead Revisited   (1945)}                       & 0.0006   \\ 
	\multicolumn{1}{c}{Ernest Hemingway -- The Old Man and the   Sea (1952)}                & 0.0006   \\ \hline
	\end{tabular*}
	\label{tab:pagerank}
\end{table}

For community detection, the Louvain algorithm was employed, and two prominent communities were detected, one of them entirely composed of literary works, the other comprising political, religious, philosophical, and scientific ones, as well as a minority of novels and poetry. The inclusion of some novels in the latter community is justifiable, demonstrating the model's ability to look past genre to detect the strong philosophical and didactic content within them. For example, significant portions of Victor Hugo's \textit{Misérables} and Jean-Jacques Rousseau's \textit{New Héloïse} discuss the authors' viewpoints on various matters, while others are not as tenable, but might suggest unexpected, subtle connections worth exploring. Additionally, the analysis revealed a powerful authorial signal: in all cases, the works of a single author were grouped within the same community. This finding serves as another validation, indicating that the model is sensitive enough to consistently capture the unique stylistic fingerprint and thematic preoccupations of an author.

\section{Conclusion and future work}

This paper explored the possibility of quantifying intertextuality with the pairwise comparison of n-gram embeddings and showed its efficacy in moving beyond the limitations of exact lexical matching and producing a global intertextuality score that aligns with human interpretation, in terms of diction, style, genres and themes. It also applied the method to a medium-sized corpus of texts of all types, proving its scalability, and subsequently performed network analysis, the results of which further validated the model.

The strength of this novel quantitative model lies in its simplicity, interpretability and computational efficiency. Its underlying principles are easy to grasp, and, requiring only a trained word embedding model and consisting of four steps---extracting n-grams, embedding them into a vector space, computing their pairwise similarities and averaging the results, it is straightforward to implement. The method is also highly parallelisable, making it feasible to apply to large corpora. The use of n-grams allows for a more specific capture of intertextual references than single words, while the embedding-based approach overcomes the limitations of exact lexical matching.

The two experiments---the validation study on four texts and the network analysis on a larger corpus---both strongly demonstrated the effectiveness of the proposed method across different analytical scales. The first experiment, a focused comparison of four texts, confirmed that the model's quantitative scores align with the nuanced, qualitative human understanding derived from close reading. In this micro-level analysis, the method successfully captured the degrees of stylistic, thematic, and authorial connections. The second experiment, conducted on 267 texts, showcased the method's utility as a tool for distant reading. On this macro-level, the network analysis revealed large-scale patterns of influence and community structures, thereby confirming the method's capacity of generating meaningful insights at scale. The success of the model at both the close- and distant-reading levels underlines its versatility as an aid to computational literary studies.

However, it is conceded that the proposed method has certain limitations. Meaningless pairwise comparisons are only filtered out after the computation of similarity, while ideally, such comparisons could be filtered out before the similarity computation to reduce computational cost. Moreover, some phrases are more evocative than others, and thus should be assigned more weights when we are calculating intertextuality. We can borrow some ideas from traditional keyword extraction methods, such as TF-IDF, to devise a quantitative way to represent evocativeness, but this is left for future work.

\appendix

\section{Rules used for normalising spelling}

There are 19 rules dealing with early modern English (`\$' denotes the end of a word):

\begin{enumerate}
	\item Long s (ſ) \textrightarrow s
	\item æ \textrightarrow ae, œ \textrightarrow oe
	\item u $\xleftrightarrow{}$ v
	\item i $\xleftrightarrow{}$ j
	\item y $\xleftrightarrow{}$ i
	\item vv \textrightarrow w
	\item ck \textrightarrow c/k
	\item ph \textrightarrow f
	\item t\$ \textrightarrow ed
	\item (')d\$ \textrightarrow ed
	\item e\$ \textrightarrow REMOVE
	\item es\$ \textrightarrow s
	\item ie\$ \textrightarrow y
	\item ee \textrightarrow ea
	\item ement\$ \textrightarrow ment
	\item ence\$ \textrightarrow ance
	\item ll \textrightarrow l
	\item ae \textrightarrow e
	\item oe \textrightarrow e
\end{enumerate}

Seven additional rules convert Americanisms into their British equivalents:

\begin{enumerate}
	\item or\$ \textrightarrow our
	\item re\$ \textrightarrow er
	\item se\$ \textrightarrow ce
	\item yze\$ \textrightarrow yse, and conjugated forms
	\item led\$ \textrightarrow lled, ling\$ \textrightarrow lling, ler\$ \textrightarrow ller, lor\$ \textrightarrow llor
	\item e \textrightarrow ae/oe
	\item og\$ \textrightarrow ogue
\end{enumerate}

\section{Data availability}
\label{sec:data}

The code and the best-performing embedding model have been released to GitHub \citep{xing_intertextuality_2025}.

\bibliography{reference}

\end{document}